# Deep learning for 3D point cloud processing – from approaches, tasks to its implications on urban and environmental applications

Zhenxin Zhang[1], Zhihua Xu[2], Yuwei Cao[3], Ningli Xu[4,5,6], Shuye Wang[1], Shen'ao Cui[1], Zhen Li[1], Rongjun Qin[4,5,6,7]*

1. Key Laboratory of 3D Information Acquisition and Application, MOE, Capital Normal University, Beijing, and College of Resource Environment and Tourism, Capital Normal University, Beijing 100048, China
2. College of Geoscience and Surveying Engineering, China University of Mining and Technology (Beijing), Beijing 100083, China
3. Department of Forest Resources Management, University of British Columbia, 2424 Main Mall, Vancouver, British Columbia, V6T 1Z4, Canada
4. Geospatial Data Analytics Laboratory, The Ohio State University, Columbus, OH 43210, USA
5. Department of Civil, Environmental and Geodetic Engineering, The Ohio State University, Columbus, OH 43210, USA
6. Department of Electrical and Computer Engineering, The Ohio State University, Columbus, OH 43210, USA.
7. Translational Data Analytics Institute, The Ohio State University, Columbus, OH 43210, USA.

*Corresponding Author

**Abstract:** Point cloud processing as a fundamental task in the field of geomatics and computer vision, has been supporting tasks and applications at different scales from air to ground, including mapping, environmental monitoring, urban/tree structure modeling, automated driving, robotics, disaster responses etc. Due to the rapid development of deep learning, point cloud processing algorithms have nowadays been almost explicitly dominated by learning-based approaches, most of which are yet transitioned into real-world practices. Existing surveys primarily focus on the ever-updating network architecture to accommodate unordered point clouds, largely ignoring their practical values in typical point cloud processing applications, in which extra-large volume of data, diverse scene contents, varying point density, data modality need to be considered. In this paper, we provide a meta review on deep learning approaches and datasets that cover a selection of critical tasks of point cloud processing in use such as scene completion, registration, semantic segmentation, and modeling. By reviewing a broad range of urban and environmental applications these tasks can support, we identify gaps to be closed as these methods transformed into applications and draw concluding remarks in both the algorithmic and practical aspects of the surveyed methods.

## 1. Introduction

Point clouds as a fundamental source of 3D data, have been in use before the development of digital computers [1], [2]. Modern sensors and algorithmic development in 3D data acquisition and processing have pushed its success to an unprecedented level such that, even personal smartphones hold sensors and software applications (Apps in short) [3], [4] that possess the capability in generating accurate 3D point clouds. Professional and consumer-grade 3D systems, such as LiDAR, RGB-D (D standards for depth) sensors, metric cameras, photogrammetric processing systems, Radar systems etc., can nowadays be mounted through platforms ranging from air to ground, with collection scale as small as biometric fingerprints to country-level 3D point clouds [5] for city-wide urban structures, driving various real-world geospatial applications in autonomous driving, navigation, mapping, forestry, and mining industry, etc.

3D point clouds generally have three types of fundamental functions in the field of geomatics and computer vision, namely, 1) measurements/visualization; 2) recognition and segmentation for automated systems; 3) geometric modeling of objects. However, turning the raw 3D point cloud input into products serving for these purposes often requires robust algorithms and processes. For example, to complete the 3D scanning of an entire scene or object, it requires *registration* of point clouds produced from multiple partial scans, and sometimes these 3D scans can be from different sensors. Moreover, when points are absence in scans due to occlusion or the sparsity of the point clouds, *scene completion* are often needed to present a continuous scene representation for downstream tasks [6]. In mapping and autonomous driving, point clouds are acquired from LiDAR and photogrammetry, and further processing is needed for modeling, localization, thematic mapping, facility management, as well as real-time obstacle avoidance, route planning etc. [7]. **Semantic segmentation and classification** interpret the object and instance-level categories for each point in a point cloud and produce a machine interpretable format. Existing engineering practices have well adapted point clouds as a product from various sensors, such as those on sensor integration in use by drive-less cars, laser scanning for bridge inspection, Building Information Modeling (BIM), etc. However, the accuracy or automation of these processes remain the major challenges. For example, point clouds have been well used for obstacle avoidance in autonomous driving, while its point density can still be limited to capture thinner objects such as poles, needing to be densified through **point cloud/depth completion**. Moreover, errors of the semantic segmentation, such as classifying non-ground objects as ground, may cause serious safety issues [8]. Furthermore, in the geospatial industry, 2D/3D asset creation, monitoring and management are still subject to intensive and manual processes, for example, creating level-of-detail (LoD) models requires reconstructing the topology of intricate building components, and are still subject to intensive manual editing, and BIM reconstruction is almost a completely manual process, since the complex indoor geometry can be hardly vectorized by existing methods [9].

There have been a plethora of review focusing on individual topics with methodological details and taxonomies of various point cloud processing tasks, such as on shape/object and segmentation [10], [11], semantic segmentation [12]. The use of Transformers in point cloud data processing [13], [14], and point cloud registration [15] etc. have provided detailed review of the algorithmic/modeling components, mostly related to various deep learning (DL) architecture. However, critical synthesis is missing about the connection of these models/algorithms to point cloud processing tasks in practice, as well as the possible roles of different learning models in these tasks as compared to traditional point cloud processing methods. For example, for algorithms for processing large volumes of data, critical synthesis is needed to assess their computation efficiency, ability to handle large scale differences, and their potential to support parallel and memory-efficient computation for billion-points processing tasks. Similarly, considerations on the use of high-capacity models (deep models) with large number of training samples, versus, the use of relatively shallower and less transferable models with fewer samples are yet to be discussed in practical applications when it comes to operational possibility and effectiveness. However, deep models are general data/pattern learners, thus, different tasks, such as registration, semantic segmentation, and modeling may share similar deep learning architecture, the role of which can be summarized to minimize the conceptual duplications in point cloud processing approaches. In this paper, we close these gaps by providing a meta-review with specific focuses on the application aspects of point cloud processing algorithms and their current use in the practice. Instead of enumerating the already well-known DL architectures for handling point cloud or their derived data, we draw the line connecting algorithms to their respective applications with focuses on a comprehensive overview of both the algorithms and the tasks they support, with a summary on basic learning models and architectures used cross-tasks to minimize the redundancy of methodology overview.

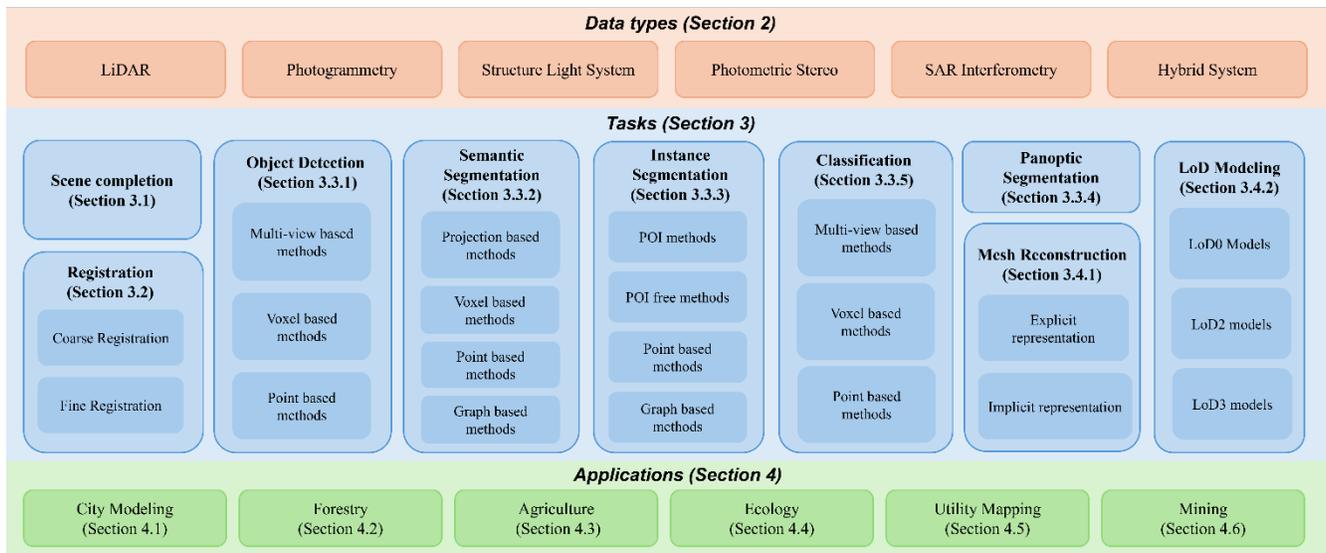

*Figure 1. An overview of topics surveyed in this review. Section 2 mainly discussed the various types of point clouds. Section 3 contains the main tasks of point clouds in photogrammetry community. Section 4 discussed the main down-streaming applications.*

In this review, we assume the input point clouds are source agnostic, thus sensor specific processes, such as radiometric calibration for LiDAR, and photogrammetric/geo-referencing process for photogrammetry etc., will not be explicitly discussed. Due to the length of the review paper, we will review four primary tasks related to point cloud processing, including: 1) scene completion; 2) point cloud registration; 3) segmentation; 4) modeling, which are widely used in their downstream applications. Since this is carried out as a meta-review, experimentation and algorithm comparison will not be performed in this work, rather, discussions on the results are reflected from existing literature. Figure 1 provides an overview of topics covered in this review encapsulating the data, tasks and the relevant applications, respectively introduced in Sections 2 to 4.

The remainder of the paper is organized as follows, Section 2 provides an overview of means to generate point cloud data to provide the context; Section 3, as one of the core sections of this paper, overviews the above-mentioned four tasks and the associated algorithms; in Section 4, we review the supported downstream applications such as in city modeling, forestry, agriculture, Ecology and Utility Mapping; Section 5 discusses approaches and applications we reviewed in this paper and Section 6 concludes this review with remarks and outlooks of deep models in point cloud processing.

## 2. An overview of point cloud data

3D point clouds can be collected or generated from a variety of different sensors and methods; primary means include: 1) laser scanning/LiDAR; 2) photogrammetry processing of images; 3) structured light systems; 4) photometric stereo/shape from shading; 5) synthetic aperture Radar (SAR) based interferometry; 6) hybrid LiDAR and image collection systems. Figure 2 and Table 1. provides a brief overview of these methods and their characteristics, which will be entailed in the following sections.

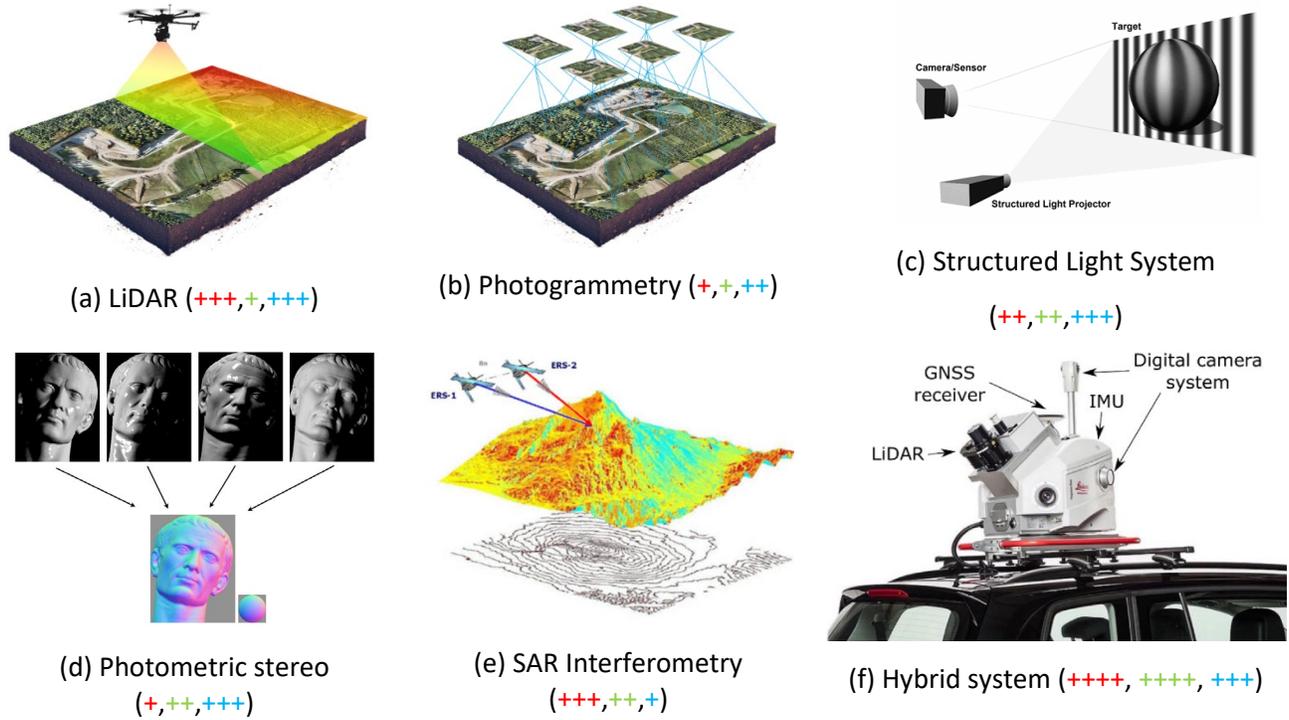

Figure 2. Overview of point cloud acquisition methods. "+" means the sensor cost, "+" pre-processing complexity, "+" means geo-reference accuracy. Photo courtesy: (a)(b) [16], (c) [17], (d) [18], (e) [19] and (f) [20].

Table 1. An overview of point cloud acquisition methods. "+" means the various levels of cost, complexity or accuracy.

|  | Sensor cost | Collection scale | Pre-processing complexity | Appearance information | Georeferencing/ Accuracy |
|---|---|---|---|---|---|
| LiDAR | +++ | Ground, Aerial | + | Intensity | +++ |
| Photogrammetry | + | Ground, Aerial, Satellite | + | RGB / Multispectral | ++ |
| Structured Light System | ++ | Ground, Local | ++ | RGB | +++ |
| Photometric stereo | + | Local, Indoor and at the object scale | ++ | N.A. | +++ |
| SAR Interferometry | +++ | Aerial, Satellite | ++ | N.A. | + |
| Hybrid system | ++++ | Ground, Aerial | ++++ | RGB/Multispectral | +++ |

***Laser scanning / LiDAR***: Laser scanner/LiDAR (Light Detection and Ranging), as a type of active sensor, determines 3D measurements using time of flight (TOF) of laser or phase shift of full wavefront [21], [22], [23]. With a rotating scanning mode, it captures thousands to millions of points per second. LiDAR sensors are available on both air and ground platforms. Typically, point clouds acquired through LiDAR come with highly accurate positional information of objects, which are often coupled with accurate Global Positioning System

(GPS)/ Inertial Measurement Unit (IMU) positional sensors to obtain high absolute accuracy. One of its major advantages is that LiDAR-based point clouds can directly provide 3D measurements, the accuracy of which is independent of the shape of the objects and scene complexity, making it an ideal instrument for data collection in complex scenes. Given its high geometric accuracy, it is a favorable source for many modeling tasks such as for buildings and trees. Moreover, due to its capability of receiving multiple returns, in contrast to other 3D collection methods, it can penetrate sparse obstructions such as vegetated layers, which is particularly effective in separating canopy and the underneath terrains [24]. However, like many other optical sensors, LiDAR may fall short in reflecting surfaces such as glass windows and water surfaces. Moreover, LiDAR sensors can be rather expensive: despite there are low-cost and consumer-grade sensors for the ever-developing autonomous driving industry, high-quality, surveying-grade LiDAR sensors are still costly to be scaled for consumer-grade platforms.

***Photogrammetric point clouds:*** Photogrammetric point clouds refer to those generated from images through the photogrammetric/structure from motion process. As a type of passive sensing technique, it remains to be one of the most versatile means in platforms running from sea to space. A major characteristic of the photogrammetric point clouds is that these 3D measurements are generated from dense matching algorithms, which however, are very often scene dependent. For example, photogrammetric point clouds of an indoor environment, may contain a notable level of noises introduced both in the bundle adjustment (BA) and dense matching algorithms, due to the lack of distinctive features (plane walls) and short baseline images in confined space (poor camera networks). As a comparison, point clouds from aerial photogrammetric images are often less noisy due to their better design image collection pattern and camera networks, i.e., more convergent images, minimal image distortions (fronto-parallel configuration) and rich texture. As a typical optical sensor that obtain 3D measurement by triangulation, point clouds of photogrammetry generally suffer from reflecting, transparent, complex surfaces, and texture-less regions [25], [26]. However, as a low-cost and very flexible means, it is still an extremely attractive approach to generate 3D point clouds.

***Structured light system:*** Structured light systems (SLS) utilize structured light patterns cast onto an object, observe their deformed patterns, and then compute the surface of the object. There are two types of SLS [27], [28]. 1) stereo-camera based SLS and, 2) single-camera based SLS. Type 1) utilizes the similar concept as photogrammetry, which triangulates point correspondences of images to obtain 3D measurements [27]. Different from photogrammetry, SLS has active lighting components (using visible or infrared lights) to cast patterns on the object of interest, on which correspondence search algorithms utilize this coded pattern to localize accurate correspondences. Type 2) uses a single camera, while it casts dynamic lighting patterns with frequency but changing phases. The reconstruction algorithm captures the deformation of the light patterns to infer the geometry (using phase unwrapping) [28]. Typically, SLS is used for scanning smaller objects in close-range and offers higher accuracy at the millimeter and micro level. As the SLS is moving distantly from the object, the lighting patterns become degraded and therefore it will no longer work. Moreover, SLS is mostly used for modeling the geometry of the object and often requires additional captures to have texture information of the objects. Furthermore, like many other optical light-based methods, SLS may remain deficient in capturing reflective and transparent surfaces.

***Photometric Stereo:*** Photometric stereo (PS), also known as, shape from shading, is another approach to acquire 3D information from images. Different from photogrammetry, which captures multiple images of the same scene from different perspectives. PS, instead, keeps the camera still, and captures images by actively changing the direction of the lighting. By using a physics-based approach [29], PS resolves the surface normal and the albedo of the objects at the same time [30]. PS typically offers high-accuracy grided point clouds for reconstruction of fine details of small object surfaces, while it has certain limitations: first, it must work under a low-light environment that the controlled light plays the major role in illuminating the objects; second, it is generally challenging to work with reflective and transparent surfaces [31], although there exist several partial solutions [31]. Third, it requires the known lighting directions for each image, which can be demanding in general practices. Despite of these limitations, PS, does have found its unique applications in small object scanning, such as teeth, fingerprint scanning, and industrial level small objects through vision metrology [32], etc.

***SAR Interferometry:*** SAR interferometry is a technique that captures 3D information using the radar inference [33], [34]: using two passes of scan with the same frequency, the topological differences of the ground surfaces will incur phase shifts of the radar signal, thereby a phase-unwrapping process taking these phase shifts will produce 3D measurements. When the number of the passes are plenty (i.e., > 10), a similar solver but designed for multiple passes called SAR Tomography [35], can be used to obtain higher accuracy 3D point clouds. SAR-based point clouds have their unique advantage that can penetrate clouds and aerosols. SAR interferometry has the capability to provide highly accurate measurements on the geometric differences and are mostly used for deformation analysis in land subsidence assessment. However, Due to the significant speckle noises of the SAR images, the level of noises/outliers from the SAR point clouds is rather high, therefore, only sparse, and selectively reliable points are used for such analysis.

***Hybrid Sensors (optical camera + LiDAR):*** Nowadays the airborne LiDAR system typically comes with a coupled high-resolution camera, therefore, it offers the ability to acquire both accuracy LiDAR point clouds and high-resolution texture and color (including multi-spectral bands). Similar sensor suites at the ground level by mobile mapping systems (MMS) [36], [37] are also available. With appropriate sensor calibration and results fusion, colored point clouds (including intensity from the original LiDAR) from such hybrid sensors are highly accurate and provide sufficient contextual information about the scene [38]. However, due to the cost of such sensor suites and the proprietary data formatting, such data are rather limited. Therefore, deep models utilize data of this quality can be rather scarce.

## 3. An overview of point cloud processing tasks

### 3.1. Scene completion

Point clouds represent 3D environments through a collection of discrete points, offering an essential and accurate depiction of real-world objects. The development of advanced sensory technologies, such as LiDAR and structured light scanners, has significantly improved the quality and resolution of point cloud data for modeling of complex environments. Nevertheless, several challenges remain, primarily attributed to factors such as restricted viewing angles, occlusions, and limited resolution. These challenges frequently result in incomplete or partial point clouds that inadequately represent the scene or object, thereby compromising the effectiveness of downstream tasks. Advancements in deep learning have revolutionized the field of scene completion using point cloud data. In scene completion, deep learning techniques understand the scene structure from the sparse points, fill gaps and densify the points in 3D spaces, providing a more comprehensive and coherent representation of the environment.

Deep learning-based approaches utilizing neural networks have been developed to address the challenges posed by missing or incomplete 3D data. Due to the unstructured nature of point clouds, conventional convolution-based methods typically require pre-processing steps to transform point clouds into structured formats, such as voxels or 3D grids for effective network operations. For instance, Wang et al. [39] utilized edge shape prior knowledge and recovered more realistic surface details by embedding point clouds into voxel networks and learning multi-scale grid features, thereby achieving better performance in handling incomplete point cloud data. While voxelization is necessary for compatibility with standard convolutional architectures, it often results in the loss of fine local geometric information, which is critical for the precise reconstruction of complex shapes. Moreover, the choice of voxel size significantly affects both the resulting resolution and computational efficiency of the scene completion. Smaller voxel sizes allow for higher-resolution reconstructions but also substantially increase memory requirements and computational time [40]. DepthSSC leveraged monocular inputs and introduces dynamic voxel resolution adjustment via Geometric-Aware Voxelization (GAV), enhancing depth alignment and boundary precision. This strategy improved 3D semantic scene completion accuracy while maintaining efficiency, as validated on benchmarks [41].

Point-based networks, which are designed to directly process raw point cloud data while maintaining the inherent permutation invariance of point clouds (i.e. invariant to the ordering of the points). Recently, several methods have been developed to simultaneously capture both local and global features. These networks typically adopt an encoder-decoder framework, to generate deep embeddings of the point clouds for scene completion. A recent method introduced a joint skeleton-surface completion network ($S^2$-PCN), which disentangled and reconstructed both global structure and local surface details via dual decoders and attentive coupling, significantly improving the realism and accuracy of completed point clouds [42]. SCPNet (Semantic scene Completion on Point Cloud) [43] is a deep model for point cloud scene completion; By using a multi-frame model knowledge distillation strategy with post label correction (for dynamic objects and other noises), it is particularly effective to deal with point clouds containing multi-scale and dynamic objects. Direct one-step completion networks, such as [44], utilized advanced modules like point contextual transformers and external attention mechanisms to generate detailed and accurate point clouds. SOAP proposed a structure-oriented autoregressive model for point cloud completion, modeling point generation as a sequential process guided by object geometry. This method excels in preserving fine structures and continuity, outperforming previous methods on benchmarks like ShapeNet and KITTI [45]. Techniques [46] incorporating block-based contextual encoders and semantically guided decoders leverage RGB-D images to enhance scene completion, bridging the gap between 2D and 3D data representations and improving the robustness and comprehensiveness of the final output. PointSea introduced a hierarchical completion strategy via point-level and object-level priors, enabling instance-aware geometry restoration in sparse scenes. Its multi-scale object representation improved robustness and generalization, particularly under partial occlusions and cluttered environments [47].

Notably, the application of generative adversarial networks (GANs) [48], particularly in inpainting tasks, has demonstrated significant success in completing occluded structures. Scene completion generally refers to the task of reconstructing missing parts of a 3D scene, while inpainting specifically focuses on filling in smaller gaps or missing areas within the scene. These networks leverage their ability to generate new data that shares the properties of the input through a generator-discriminator framework. Specifically, the generator produces plausible outputs, while the discriminator evaluates the generated data against the real input, thereby refining the process iteratively. However, the unordered nature of point clouds introduces ambiguity, which can potentially affect both training efficiency and the quality of results. Recent developments have explored combining GANs with representation learning and multi-view supervision. SSRN-GAN, for instance, integrated semantic-aware adversarial learning with spatial-aware residual refinement, enhancing both local fidelity and global structural continuity in incomplete indoor scenes [49]. Furthermore, lightweight GAN architectures using hybrid 2D-3D distillation and depth priors had shown promising efficiency in large-scale reconstructions [50]. These frameworks either directly refine point-based geometry or infer dense 3D structures from sparse inputs, driving progress toward faster and more realistic scene completion. Cross-domain frameworks [51] employing class-conditional GAN inversion had shown promise in restoring authentic shapes from incomplete inputs, highlighting the dynamic evolution of point cloud completion technologies. These methods either directly reconstruct the geometric features of a scene or infer dense 3D shapes by analyzing the underlying spatial distribution. The goal of these approaches is to achieve faster, more detailed, and continuous 3D scene completion.

**3.2. Registration**

Point cloud registration is the process of estimating the transformation relationship (e.g., rigid, similarity, non-rigid transformation) between a pair of 3D data sets to minimize their systematic error, which is widely used in 3D mapping and change detection tasks. Existing methods can be categorized into coarse and fine registration according to the level of registration accuracy. The registration accuracy depends on different levels of information involved in the transformation parameter estimation process, where coarse registration methods usually are based on sparse correspondences of the key primitives and fine registration methods are based on the dense correspondence of the whole point clouds.

*3.2.1. Coarse registration*

Coarse registration methods usually construct sparse correspondences which are invariant under pre-defined transformation relationship, and then estimate the parameters based on the extracted correspondences. The sparsity of the correspondences makes this type of method not achieve the optimal registration accuracy.

Coarse registration estimates transformations via sparse invariant correspondences, but their inherent sparsity fundamentally limits registration precision due to insufficient geometric constraints. **Correspondence extraction** can be achieved by three steps: 1) primitive detection, 2) feature description and 3) feature matching. The primitives that are usually used for registration are key points, lines, planes and patches. Key points [52], [53], [54], [55], [56], [57] are the most widely used 3D features due to their simplicity, whereas for urban scenes and man-made objects, line-based or plane-based representation can achieve better performance [58], [59]. Learning-based key-point detection methods [53], [54], [55] received increasing attention. Li et al. [53] proposed a PointNet-like network [60], [61] to predict the key-points with the saliency uncertainties in an unsupervised way. Huang et al. [55] further solved the partial overlap case by using a KPconv-like network [62] to predict the overlap score, match ability score per point as the saliency value.

To match these extracted geometric features, a process called feature description is used to encode unique information about the features (key points, lines, planes or patches), which is usually done by extracting local characteristics from the local surface reflecting information related to 3D location, gradients and appearance (if color channels are applicable). This process has been advanced by data-driven methods replacing hand-crafted feature descriptors with learning-based descriptors. Most of the works are based on point-based features due to the simplicity of point primitive. Local-based descriptors [63], [64], [65], [66], [67], [68], [69], [70] aim to learn the general transformation-invariant features from local patches around the key-points, which are demonstrated to robust to unseen scenarios. Ao et al. [63] first applied a spatial point transformer network to map the local patches to the cylindrical space, a SO(2) equivalent representation, and then proposed a neural feature extractor based on 3D cylindrical convolutional neural networks (CNN) [71] to learn the features. Li et al. [69] improved the 3D feature description performance by introducing the multi-view rendering module that use the powerful 2D CNN to learn the 2D features from rendered views looking at the local patch and combine these features as the 3D local descriptor. Global descriptors [54], [55] used end-to-end network architecture that takes the whole point cloud as input and predict the per-point descriptors, which are yielded global features against registering partially overlapped regions. Huang et al. [55] applied a cross-attention module to detect the region of similar shape between two sets of point clouds and predict the per-point overlap score, which can instruct the later point matching on high overlap score regions.

Once features and their descriptions are extracted, correspondences can be found by finding the most similar features between two datasets. The metric measuring the similarity is then defined in a vector space (e.g., hamming distance, Euclidean distance). To reduce the outliers, this matching process can be further constrained by augmenting the distinctiveness of the correspondence through standard methods such as ratio test [72], spectral matching [73] or tuple test [74]. Recently, the learning-based methods are proposed to directly learn to construct the correspondence [54], [75], [76], [77]. These approaches offer the advantage of encoding the entire point cloud into feature representations, enabling correspondence construction at a global level rather than being limited to local regions.

**Parameter Estimation** for rigid transformation requires as few as three key-point correspondences. While sparse 3D correspondences through feature extractions often contain a large number of outliers. The outlier rates of 3D correspondences are extremely high (e.g., as high as 95%) [78]. Therefore, current research trends focus on developing methods that are robust to correspondences of high outlier rates [74], [78], [79], [80], [81], [82], [83], [84], [85]. Traditional methods follow either RANSAC-like mechanisms [86], [87], [88], [89] that randomly sample the minimal sets of correspondences to maximum the consensus set or M-estimation mechanisms [74], [80] that globally optimize the robust object functions with respect to noisy correspondences. Deep-learning based methods [83], [84], [90] aims to learn the robust features to encode contextual information to accurately and robustly describe correspondence potentials before registration. PointDSC uses a few neural network-based modules to map and group initial correspondences at the deep feature level. Correspondences from the same group are assumed to share similar geometric consistency and are tested for registration, and the solution agreeing with most correspondences is selected as the estimated solution.

*3.2.2. Fine registration*

Assuming the rough initial pose is available, the fine registration methods usually iterate between dense correspondence construction and parameter estimation until a certain criterion is fulfilled. The utilization of dense information and the iterative optimization solution makes this type of method achieve the optimal registration accuracy. There are mainly three types of methods: ICP-derived and probabilistic methods.

**ICP-derived methods.** The ICP-derived methods [91], [92], [93], [94], [95], [96], [97], [98], [99] have been popular in the literature for over thirty years due to their simplicity and effectiveness. Standard ICP iterates between constructing the correspondences by the closest point and updating the transformation parameters. Its nature of iterative optimization makes it sensitive to the initial pose and thus its convergence is not guaranteed. Recent deep-learning based methods [100], [101], [102], [103] follow the same iterative framework while the improvement is in the correspondence construction part, where they replace the method using closest point as correspondences with the neural network to improve the robustness of the constructed correspondences. Specifically, Wang et al. [100] constructed the "soft" correspondences by using point-based network, DGCNN [104] with the attention module to construct the soft map. Similarly, Li et al. [102] proposed a similarity matrix convolution to learn to select the top-ranked correspondences.

**Probabilistic methods**. The probabilistic methods [105], [106], [107], [108], [109] model the raw point clouds as samples from certain probability distributions (e.g. gaussian mixture model (GMM). Thus, the point cloud registration problem is reformulated as a problem to minimize two data distributions. Most existing methods use GMM to model the point clouds, where point clouds can be either the samples of the assumed GMM model or the GMM itself (by taking the location of each point as the centroid). The minimization can be achieved through maximal likelihood estimators (MLE) represented by GMM. Yuan et al. [110] improved the correspondence by matching each point to a GMM learnable parameters and the transformation parameters are estimated through the differentiable modules. Mei et al. [111] solved the partial overlap registration problem by introducing a point cloud classification module to identify the overlap region.

One common challenge for deep-learning-based methods is the handling of high-volume data, such as city-scale or LiDAR point clouds. Coarse registration methods are inherently suitable for managing large data volumes. End-to-end methods like PREDATOR and D3Feat process down-sampled versions of input point clouds to directly estimate sparse correspondences and transformation parameters. Descriptor-based methods, such as SpinNet and YOHO, extract sparse correspondence by identifying key primitives and corresponding local patches for 3D description. These methods offer better accuracy than end-to-end approaches because the local patches retain all the information. Fine registration methods struggle with high computation and memory consumption when dealing with large data volumes. This is because their dense correspondences are typically constructed using Kd-**trees**, which have a space complexity of O(N) and a query time complexity of O (log N). Existing methods like motion average [97] often transform large-scale point clouds into grid structures to reduce space and query complexity to constant levels.

Another challenge is the low generalizability when transitioning from training scenes to unseen scenes. Most methods trained on small-scale indoor datasets, such as 3DMatch [68] and ModelNet40 [112] , experience a significant performance drop on outdoor and large-scale datasets like odometryKITTI [113], ETH [114], and WHU-TLS [58]. Deep-learning models that focus on learning components within the entire registration pipeline, such as 3D descriptors [63], [64] and parameter estimation [83] have demonstrated better generalizability than end-to-end methods.

### 3.3. Object Detection, Segmentation and Classification

Detection, segmentation and classification of different types of objects from the point cloud is foundational for many point cloud applications. Detection is the process of identifying and localizing objects within a 3D space. This involves determining both the presence and precise location of objects in the scene. Segmentation partitions a point cloud into multiple homogeneous regions or segments, each corresponding to different parts of objects or distinct objects, with each point labeled according to its respective class. Classification assigns a category or class label to the detected objects or to the entire point cloud, utilizing features extracted during detection or segmentation to classify objects based on predefined categories. While detection, segmentation, and classification address different aspects of point cloud processing, all three tasks rely on deep learning models' ability to interpret the spatial structure and distribution of point clouds.

Semantic segmentation, instance segmentation, and panoptic segmentation are three sub-topics of point cloud segmentation. Although each method addresses different aspects of segmentation, they share foundational similarities in their application of deep learning to interpret and classify point cloud data. **Semantic segmentation** assigns a class label to each point in the point cloud, delineating areas based on generic categories like road, building, or tree. This method focuses on understanding the environment at a broad level without distinguishing between individual objects of the same class. **Instance segmentation** goes a step further by not only classifying points but also identifying distinct objects within the same category—effectively separating multiple instances of a single class, such as different vehicles in a traffic scene. **Panoptic segmentation** combines the approaches of semantic and instance segmentation to provide a comprehensive scene understanding. It segments and recognizes every point in the scene, assigning each either a class label or an instance identifier. In all three cases, deep learning exploits the intricate spatial relationships and features within point clouds, enhancing segmentation accuracy and adaptability across diverse applications. These segmentation paradigms, while having distinctive objectives, their backbone network and unit feature extractors are similar, which can be generally categorized into 1) voxel-based methods, 2) multi-view-based methods, and 3) direct point-based methods.

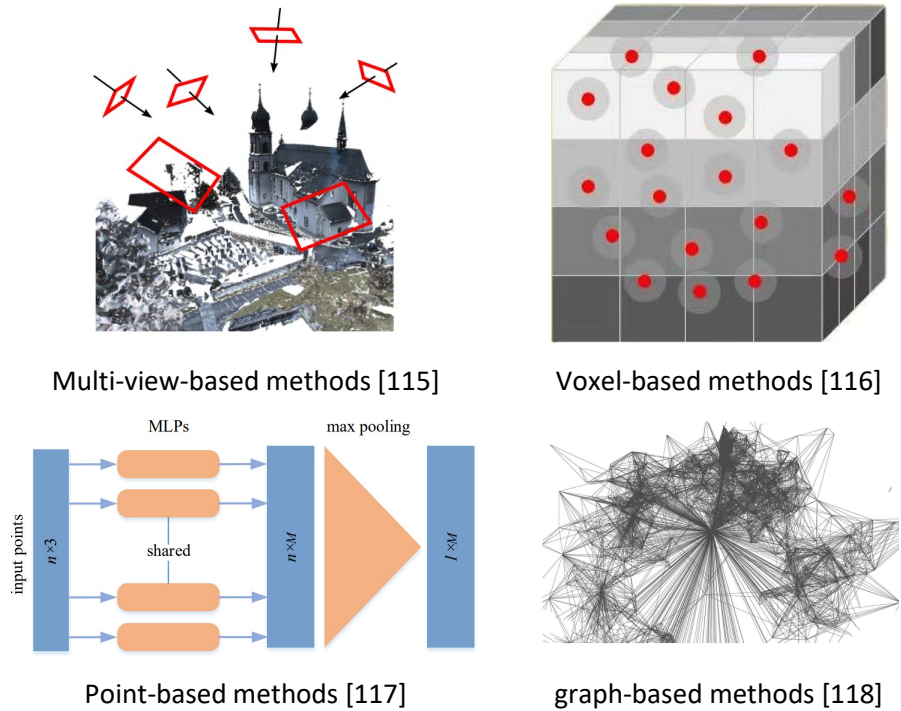

Figure 3. Different ways of feature extraction for 3D point cloud perception. More details can be found on Sec 3.3.

**Voxel-based** methods transform point clouds into a volumetric grid, akin to pixelation in images but extended to three-dimensional space. This conversion enables the application of conventional 3D CNNs that operate on regular grid structures. By voxelizing the space, these methods can harness well-established architectures and techniques from 2D image processing, adapted for 3D data. Voxel-based approaches excel in capturing local structural patterns within the grid, thereby facilitating robust feature extraction for detection, segmentation, and classification tasks.

**Multi-view-based** methods address point cloud processing by generating multiple 2D projections of the 3D space from diverse perspectives. This approach leverages the well-established capabilities of traditional 2D CNNs, applying them to each projection independently before integrating the results for a final decision. The primary advantage of multi-view approaches is their ability to utilize mature image processing techniques to extract features from complex 3D structures represented in a computationally efficient 2D format. Each projection captures distinct geometric aspects of the object, and when aggregated, these views provide a comprehensive feature set that enhances the model's understanding of the 3D shape.

**Direct point-based** methods, such as PointNet [60], PointNet++ [119], directly process unstructured raw 3D points for feature extraction. The key to these approaches lies in the design of novel network architectures that effectively handle unstructured 3D points through neighborhood graphs and spatial encodings [120]. These methods are inherently invariant to the order of points, providing flexibility in practical applications, such as Indoor scene segmentation [121] and autonomous driving perception [122]. This makes them particularly suitable for segmentation tasks where input 3D points exhibit diverse structures, densities, and complexities.

Despite their procedural differences, these approaches aim to extract representative features or embeddings from 3D point clouds for subsequent tasks. Essentially, they transform irregular and unordered 3D data into a structured format that deep learning models can effectively process. Each of these three methods develops specific deep learning models to capture the spatial relationships and intrinsic properties of the data. The following subsections on methods for different tasks follow this categorization.

*3.3.1. Object Detection*

Deep learning has significantly advanced the field of object detection using point cloud data. Given a set of dense point clouds, the objective is to predict object categories and their spatial boundaries by generating 3D bounding boxes of the detected objects. Following the taxonomy described above, the relevant approaches for object detection are described below.

**Voxel-based methods** project point cloud data into voxel grids, subsequently extracting features from individual and clusters of voxels as fundamental units. VoxelNet [123] efficiently transforms raw point clouds into structured voxel grids by employing specialized feature encoding layers that integrate the intrinsic properties of individual points with the overarching structure provided by 3D voxelization [124]. To enhance object detection robustness, researchers have explored multimodal data integration, such as combining image and point cloud features within the voxel grid framework [125]. Advanced techniques, including sparse 3D convolutions and submanifold sparse convolutions, further embed deeper and more unique features into the voxel grid, thereby improving computational efficiency and reducing memory consumption. Although voxel-based methods are straightforward for feature extraction, their scalability has traditionally been constrained by the high computation and memory demands. However, recent advancements, such as the incorporation of anchor-free detection mechanisms [126] within voxel feature extraction networks [127], have significantly improved scalability. Beyond efficiency and accuracy, recent studies have also highlighted the robustness challenges of voxel-based point cloud object detectors under adversarial settings [128]. These developments enable voxel-based methods to handle large-scale scenes more effectively, maintaining their relevance in challenging object detection applications.

**Multi-view-based methods** utilize perspective views to project the spatial layout of detected objects. These methods uniquely combine point cloud attributes with 2D image data to enrich feature extraction, harnessing both the depth, texture and context information provided by 2D imagery. The idea is to make use of 2D convolutional networks designed for bounding box prediction [129], while it may sometimes struggle with depth errors caused by projection (e.g., z-buffering ambiguities and positional errors). Thus, researchers proposed to address these limitations [130] by integrating additional data layers such as high-definition maps. These maps provide geometric and semantic contexts to improve the detection, for example, road masks can be used to improve the positional accuracy of detected objects. Furthermore, methods like PointPillars [131], which utilize deep learning frameworks, such as PointNet, to extract pillar-like features directly from point clouds. By projecting these detected 3D objects to 2D view, this technique essentially serves as 2D region proposal generator, significantly boosting the processing speed. Other enhancements have focused on refining the detection process through optimized class grouping and advanced sampling techniques [132], for example, using masked autoencoders and multi-pixel fusion techniques to achieve more effective multi-view integration of 3D point cloud and 2D images [133] by leveraging the strengths of each data type to achieve a more nuanced and accurate detection capability.

**Point-based methods** enhance feature extraction from unstructured point clouds, which typically utilize techniques such as fixed-radius/bounding-box and nearest-neighborhood approaches to aggregate information for efficient and flexible feature extraction and embedding. Early implementations selected seed points from point clouds, then leveraged these points not only for feature extraction but also as a foundation for predicting 3D bounding boxes, thereby integrating feature embedding with proposal generation for object detection. This was achieved through an advanced voting mechanism, where multiple seed points contributed to refining the estimated object locations and classifications within the scene [134]. However, these initial models often struggled to capture global contextual information. To address this limitation, subsequent multi-modal methods incorporated images as additional inputs. This multi-modal strategy significantly enhanced the capability of point-based methods [135] to extract both global and detailed information, especially in complex scenarios [136]. Further innovations have specifically refined point-based detection techniques, improved feature extraction and point cloud processing algorithms. For example, per-pixel key-point voting mechanisms that leverage RGB data were developed to enhance object detection in indoor environments. By integrating these localized key-points with a broader scene-level perspective, this approach enables more accurate instance-level localization within dense point clouds [137]. Graph convolutional networks (GCNs) are also a key approach in point-based frameworks, which capture the complex spatial relationships between individual points. By refining these relationships, GCNs optimize spatial reasoning to generate more accurate bounding box predictions [138]. This addition marks a shift from earlier methods focused on local feature extraction to understanding the global context, further enhancing the model's interpretation ability. Moreover, the exploration of geometric properties through novel spatial loss functions marked another significant improvement in point-based methods. Building on the self-attention mechanisms initially developed for sequential data (such as text), transformer architectures have now been adapted to process point clouds by treating points—or groupings of points—as "tokens." This design have been integrated into point-based systems [139], significantly boosting performance by enhancing the model's capacity to discern intricate patterns in large-scale point clouds. Overall, point-based methods in 3D object detection have advanced from rudimentary feature extraction techniques to sophisticated systems that integrate diverse data types and cutting-edge computational models, these advancements have significantly enhanced both detection accuracy and efficiency. These developments underscore an ongoing commitment to maximizing the potential of point cloud data for a deeper understanding and more effective interaction with three-dimensional environments.

*3.3.2. Semantic Segmentation*

As mentioned below in Section 3.3.1, semantic segmentation follows a similar paradigm at the unit level for feature extraction. However, unlike object detection, which focuses on bounding box localization and instance-level identification, semantic segmentation assigns class labels to each point or region in the data. This distinction underscores the unique challenges and methodologies required for segmentation tasks. **Projection-based** segmentation methods transform 3D point clouds into 2D images or 360-degree panoramic views, leveraging traditional image segmentation techniques. This method simplifies complex 3D data, making it accessible to powerful 2D CNNs. **Voxel-based** segmentation methods divide 3D point clouds into voxels, assigning semantic labels based on the contained points. This method captures fine-grained spatial information, ideal for dense data and volumetric representation, but can be computationally expensive due to high memory requirements for large number of voxel grids. **Point-based** segmentation methods process raw point clouds directly, focusing on point relationships and leveraging advanced neural networks. It excels at capturing local structures, making it ideal for applications in autonomous driving and robotic applications, but may suffer from low efficiency in large-volume point clouds. **Graph-based** segmentation methods model the point cloud as a graph, where points serve as nodes interconnected based on their spatial or feature similarities. This method excels in capturing relational information between points, making it particularly effective for intricate and densely connected structures.

*3.3.2.1. Projection-based methods*

Projection-based methods render a number of 2D images at different perspectives from the point clouds, which are then processed by image-based semantic segmentation algorithms for fusion. For instance, Lawin et al. [140] analyzed 3D point clouds of multi-angle 2D planar projections using fully convolutional networks, merging scores from different views to obtain semantic information of points. Boulch et al. [141] subsequently employed 2D segmentation networks to perform pixel-level labeling on RGB-D data acquired from multiple cameras, and further integrated the predicted scores through residual correction techniques. Tatarchenko et al. [142] projected local point sets onto tangent planes to preserve the spatial relationships and positional information of points, attempting to address the issue of information loss. A recent study by Inadomi et al. [143] on large-scale bridge scenes further confirmed the potential of projection pipelines. Their two-stage approach combines a transparency-suppression filter, Axial-DeepLab backbones [144], and view-specific training. Despite these 2D semantic segmentation methods being developed to segment pictured 3D objects, their performances vary greatly with respect to viewpoints, occlusions, and projection algorithms.

*3.3.2.2. Voxel-based approaches*

The voxel-based methods map point cloud data onto a 3D voxel grid and segments the voxels using a 3D CNN. As compared to multi-view-based approaches, voxelization better preserves the inherent three-dimensional features of the point cloud. For example, Mei et al. [145] enhanced segmentation accuracy by predicting coarse voxel-level semantic labels and assigning these labels to all points within the voxels. Additionally, fine-level labels are further predicted and refined to improve segmentation accuracy, ensuring a more detailed and accurate semantic understanding of the 3D point clouds; Tchapmi et al. [146] introduced trilinear interpolation to improve spatial consistency, although this method preserves the spatial features of point clouds, it incurs substantial memory usage particularly when handling large-scale or high-density point clouds. In contrast, Park et al. [147] tackled the issue of real-time performance degradation due to high voxel resolution by proposing a fast semantic segmentation model using point convolutions and three-dimensional sparse convolutions, thereby mitigating the computational burden associated with high resolution while maintaining performance across different voxel resolutions. Ibrahim et al. [148] investigated the challenges posed by the sparsity and uneven density of points in outdoor point clouds when applying voxel-based feature extraction. To effectively establish object-centric feature models in point cloud data, they introduced a technique based on slot attention transformers, which excels particularly in semantic segmentation for rare object categories. Focusing on security, Wu et al. [128] systematically assessed the adversarial vulnerability of voxel-based detectors and proposed a gradient-based sparse voxel attack (GSVA) that perturbs only 3.5 % of voxels yet induces substantial mAP drops across KITTI [149], nuScenes [150] and Waymo [151], highlighting a new robustness challenge for voxel representations. However, these voxel-based approaches mentioned inevitably lead to some loss of original features. Additionally, voxelization requires a priori knowledge to determine the appropriate voxel size; low resolution data can result in loss of detailed 3D point cloud features, while high resolution can introduce significant computational and memory overhead.

*3.3.2.3. Point-based approach*

Point-based methods play a pivotal role in point cloud data processing. Qi et al. [60] introduced PointNet, which leverages multi-layer perceptron (MLP) to independently extract features from each point, achieving robust point cloud classification and segmentation results. Although PointNet excelled in global feature aggregation, it fell short in capturing local geometric details[152]. To address this limitation, Qi et al. [60] subsequently developed PointNet++, incorporating regional feature extraction via multi-scale and multi-resolution approaches to handle non-uniformity and density variations in point clouds. Jiang et al. [153] further enhanced PointNet++ by integrating a direction-encoded Scale-invariant Feature Transform (SIFT) module, thereby extracting and aggregating directional information of points and improving the model's perception of local structures.

Wu et al. [154] pioneered dynamic filters in PointConv, establishing transformation and permutation invariance for non-uniform point clouds through kernel density estimation, which laid the foundation for continuous-space convolution in point cloud processing. Building upon this, Thomas et al. [62] introduced Kernel Point Convolution (KPConv), extending PointConv's geometric adaptability by computing weights between points and learnable "kernel points." This innovation enabled dynamic kernel deformation, significantly enhancing flexibility in capturing irregular geometric structures — a critical advancement over static filters. Parallel developments focused on local relational modeling: Liu et al. [155] proposed Relation-Shape Convolutional Neural Network ( RSConv), which explicitly encoded high-dimensional geometric relationships within local subsets, while Hu et al. [156] optimized efficiency via RandLA-Net's random sampling and local feature aggregation. Notably, RandLA-Net's Local Spatial Encoding (LocSE) module can be viewed as a lightweight realization of RSConv's relational encoding, achieving a balance between computational efficiency (via sampling) and geometric preservation (via aggregation). Besides, Xu et al. [157] later enhanced this paradigm with Position Adaptive Convolution (PAConv), which abstracted KPConv's spatial weighting into dynamically assembled matrices. By decoupling geometric correlations from weight generation, PAConv provided greater network flexibility while retaining the interpretability of kernel-based approaches. The latest evolution comes from Yin et al. [158] with Double attention and Consistent constraints Network(DCNet), which synthesizes these advancements: It inherited RandLA-Net's efficient sampling framework, incorporated PAConv's dynamic weight philosophy through dual attention mechanisms, and addressed feature degradation (a limitation in PointConv-based methods) via cross-layer consistency constraints. This progression—from foundational invariance (PointConv) to adaptive kernels (KPConv), relational encoding (RSConv/RandLA-Net), dynamic weight optimization (PAConv), and holistic feature preservation (DCNet)—demonstrates a clear trajectory: advancing point cloud understanding by iteratively unifying geometric sensitivity, computational efficiency, and structural robustness.

*3.3.2.4. Graph-based approach*

Graph-based methods represent point cloud data as a graph by constructing it dynamically based on spatial proximity [156], feature similarity [159] etc., where nodes correspond to points and edges represent connections. Betsas et al. [137] presented a detailed survey that proposes a unified taxonomy and compiles an extensive repository of over four hundred deep-learning–based 3D semantic-segmentation methods and datasets, a standardized benchmark for comparing graph-based and other architectures. These methods leverage graph convolutional networks or graph attention networks for segmentation tasks. Landrieu and Simonovsky [160] pioneered the super-point graphs approach, employing directed graphs to capture contextual relationships within point clouds, thereby providing a novel perspective on semantic segmentation. Subsequently, Wang et al. [161] introduced Graph Attention Convolution (GAC), which dynamically learns and selectively captures more relevant features, albeit at the cost of increased network complexity and parameters. To mitigate these issues of complexity and feature redundancy, Zhang et al. [162] proposed the Linked Dynamic Graph CNN (LDGCNN). This method enhances network performance and reduces model size by concatenating geometric features from different dynamic graphs and utilizing MLP. Moreover, Ma et al. [163] addressed the challenge of extracting high-dimensional global semantic features by introducing the Point Global Context Reasoning module, which captures global relevance through a channel graph [163]. Most recently, Chen et al. [164] developed the Feature Graph Convolution Network with Attentive Fusion (FGC-AFNet), which improves local feature extraction and effectively balances the processing of global and local information by constructing graphs between centroid points and their neighbors. Overall, these graph-based approaches highlight the effectiveness of modeling geometric relationships through dynamic graph construction, attention mechanisms, and multi-scale feature fusion in tackling the structural complexity and semantic dependencies inherent in point cloud segmentation tasks.

*3.3.3. Instance Segmentation*

Instance segmentation of point clouds based on deep learning builds upon semantic segmentation by not only assigning semantic labels to each point but also distinguishing between individual instances within the same class, introducing a finer granularity to the segmentation process. This refinement requires additional key components, including proposal generation, which identifies regions likely containing individual instances, and internal segmentation, which assigns points within those regions to specific instances. Instance segmentation is mainly divided into two methods based on region proposal-based methods and region proposal-free methods.

*3.3.3.1. Region proposal-based methods*
Region proposal-based point cloud instance segmentation methods achieve the segmentation and recognition of distinct instances in point cloud data through the generation of candidate regions, followed by their classification and segmentation. These methods typically consist of three primary stages: candidate region generation, feature extraction, and instance segmentation. While these approaches yield superior segmentation results, they are often computationally intensive and time-consuming due to multi-stage training processes and the necessity to process many candidate regions.

The 3D Semantic Instance Segmentation of RGB-D Scans (3D-SIS) proposed by Hou et al. [165] is a typical example of a proposal-based approach, which effectively leveraged high-resolution RGB inputs by integrating 2D images with 3D meshes. Moreover, Yi et al. [166] introduced the Generative Shape Proposal Network (GSPN) model, which mitigated the generation of blind boxes through optimization of the bounding box network, albeit at the cost of a complex structure and high computational requirements. Yang et al. [167] further advanced the field with the Learning object bounding boxes for 3D instance segmentation on point clouds (3D-BoNet) , which generates candidate bounding boxes using global features; however, it falls short in capturing detailed instance-specific features. Engelmann et al. [168] enhanced segmentation precision via the multi-proposal aggregation for 3d semantic instance segmentation (3D-MPA), utilizing sparse convolutions, though the network's complexity posed challenges for optimization. Lu et al. [169] presented the Box-Supervised Simulation-assisted Mean Teacher for 3D Instance Segmentation (BSNet), which incorporates a novel simulation-assisted transformer for pseudo-labeling, providing valuable prior knowledge for instance segmentation and holding the potential to further enhance segmentation outcomes.

These proposal-based methods are key components of instance segmentation. Despite its progressive improvement over the past few years, there are still several challenges. One major challenge is that these methods cannot effectively distinguish adjacent or overlapping instances with similar features, causing false instance segments consisting of multiple objects. Moreover, proposal-based methods often face scalability issues due to their two-stage architecture (e.g., region proposal generation followed by per-proposal processing), which introduces redundant computations and high memory consumption when handling dense 3D point clouds with millions of data points. Future improvements should concentrate on enhancing feature extraction, optimizing computational efficiency, and addressing the segmentation of dense or closely spaced instances.

*3.1.1.1. Region proposal-free methods*
The region proposal-free point cloud instance segmentation method adopts a distinct strategy by directly predicting points within class instances without relying on candidate region generation. It segments and instantiates point cloud data through clustering each instance within a structured feature space. This approach obviates the need for computationally expensive candidate region proposals, thereby providing a more streamlined and efficient alternative, for instance segmentation. By concentrating on point-level predictions and leveraging clustering in the feature space, this proposal-free method addresses some of the limitations inherent in proposal-based methods.

Wang et al. [170] introduced the Similarity Group Proposal Network (SGPN), which leverages the PointNet++ backbone to extract both global and local features from point clouds. This method effectively suppresses redundant information via a three-branch prediction mechanism but suffers from high computational complexity and memory consumption, making it unsuitable for large-scale data processing. To address these limitations, Liu et al. [171] developed the Multi-scale Affinity with Sparse Convolution (MASC) , which improves segmentation efficiency by integrating U-Net with sparse convolutions. However, the feature extraction process does not fully explore multi-scale features, limiting its effectiveness in complex scenarios. Sun et al. [172]

enhanced the Similarity Group Proposal Network (SGPN) by incorporating boundary convolutions and threshold graph learning, achieving automatic threshold optimization. Lahoud et al. [173] proposed the Multi-Task Metric Learning (MTML), which adapts to complex 3D scenes by learning high-dimensional embedding features and optimizing instance label predictions. In 2021, Zhang and Wonka [174] introduced a probabilistic embedding approach that simplifies the instance segmentation process while enhancing granularity. Meanwhile, Chen et al. [175] introduced the Hierarchical Aggregation mechanism (HAIS), which mitigates over-segmentation or under-segmentation issues by gradually generating instance proposals. Recently, Zhao et al. [176] addressed the challenge of segmenting adjacent objects with similar semantic labels using the Point-Wise binarization mechanism (PBNet). PBNet effectively manages offset point distribution issues in instance segmentation by binarizing points and clustering them separately. These studies collectively demonstrate that the precision and efficiency of point cloud instance segmentation are being progressively improved through continuous algorithmic and structural optimizations.

These proposal-free methods streamline the segmentation pipeline and reduce computational complexity, which underscore the significance of novel architectures and algorithms in improving feature extraction, managing complex 3D scenes, and tackling challenges such as over-segmentation and the segmentation of adjacent objects.

*3.3.4. Panoptic Segmentation*

Panoptic segmentation for point clouds integrates semantic and instance segmentation, assigning each point a semantic category as well as an instance ID when applicable. This approach effectively bridges the gap between point-level (Section 3.3.2) and object-level understanding (Section 3.3.3). While semantic segmentation focuses on classifying every point into categories, and instance segmentation identifies distinct object instances, the panoptic segmentation unifies these two tasks within a single framework, enabling a more comprehensive scene interpretation that addresses both semantic and instance-level challenges simultaneously. The key distinction of panoptic segmentation from its constituent tasks lies in its holistic approach. Unlike semantic segmentation, which does not differentiate between multiple instances of the same class, panoptic segmentation explicitly labels each instance. Additionally, it extends instance segmentation by incorporating information about non-instance regions (e.g., background or amorphous areas). This dual-labeling system significantly enhances the generalization performance and accuracy of segmentation tasks, especially in complex scenes with overlapping or ambiguous objects.

Pham et al. [177] introduced the Multi-task pointwise Network (MT-PNet), which employs a multi-task network to jointly predict semantic categories and instance embeddings. By refining outputs through a Multi-Value Conditional Random Field (MVCRF), this approach underscores the value of integrating probabilistic models for segmentation tasks. Similarly, Wang et al. [178] developed the Associatively Segmenting Instances and Semantics(ASIS), focusing on learning semantic-aware point-level instance embeddings to effectively integrate semantic and instance features, thereby enhancing segmentation performance. Expanding on these ideas, Zhao et al. [179] proposed the Joint instance and semantic segmentation (JSNet), utilizing PointNet++ and PointConv as backbones to extract rich geometric and spatial features. However, the high memory demands of PointConv highlight the trade-offs associated with complex architectures. Building on this foundation, Chen et al. [180] introduced a joint semantic and instance two-branch pipeline (JSPNet), addressing conflicts between semantic and instance segmentation via innovations such as feature self-similarity and cross-task probability learning, thereby optimizing shared features for both tasks. Sohail et al. [181] present the first survey of deep transfer learning and domain adaptation for 3D point clouds, summarizing datasets, metrics, and transfer strategies across major tasks. Collectively, these methods illustrate a progression in network design that emphasize shared backbones, embedding refinement, and cross-task interaction to unify semantic and instance segmentation. This evolution sets a robust foundation for panoptic segmentation by addressing both pixel-level and object-level understanding within a cohesive framework. These advancements demonstrate a clear trajectory toward panoptic segmentation by integrating semantic and instance segmentation tasks. The shared backbone, feature refinement mechanisms, and cross-task interaction strategies facilitate better integration of semantic context to improve instance prediction, capture complex geometric and spatial features, and resolve task conflicts, contributing to a unified panoramic segmentation framework with higher accuracy and more complete scene understanding.

### 3.4. Geometric Modeling

There is a great demand for converting point clouds to explicit high-level structure representations that help recognize and understand the shapes [182]. Geometric modeling is the subfield of computer graphics concerning mathematical representation of 3D object shapes, which encompasses a diverse set of approaches, each offering distinct advantages and limitations in representing object complexity. Two prevalent methods are mesh reconstruction and Level of Detail (LOD) modeling. Mesh reconstruction focuses on creating polygonal meshes that represent the surface of an object, often utilizing data sources like point clouds. It finds applications in areas like 3D printing and cultural heritage preservation, where capturing intricate object details is crucial. In contrast, LOD modeling involves creating multiple representations of the same object with varying levels of detail, which allows for efficient usage in applications like real-time video games or large-scale cityscapes. This section delves deeper into these two established methods: mesh reconstruction and LOD modeling. There have been more recent work using field-based methods, such as Neural Radiance Fields (NeRF) [183] and 3D Gaussian Splatting (3DGS) [184], [185], [186], however, since these field-based methods are still largely used for visualization rather than shape/geometric modeling, we will refrain from discussing these methods under this section.

*3.4.1. Mesh reconstruction*

Recent advancements in deep learning have revolutionized the field of mesh reconstruction from point clouds. Point clouds represent 3D objects as a collection of discrete points. Mesh reconstruction takes these points and generates a more detailed representation of the object's surface as a mesh. Deep learning approaches excel in this task and can be broadly categorized into explicit and implicit methods. Understanding the strengths and weaknesses of these two categories is crucial when choosing a deep learning method for mesh reconstruction.

*3.4.1.1. Explicit representation*

Inspired by classic algorithms such as Delaunay triangulation-based surface reconstruction methods [187], explicit methods directly predict the final mesh structure, including its vertices, edges, and faces [188]. Deep learning surpasses the limitations of traditional methods when it comes to intricate surface details. Its ability to learn complex relationships from data allows it to reconstruct surfaces with higher fidelity, particularly for objects with complex shapes. For instance, Nash et al. [189] first generated mesh vertices from lowest to highest on the vertical axis, then quantized these continuous vertex positions into discrete bins for likelihood calculation, and finally generated polygon faces. Dai et al. [190] employed a MLP to generate a set of points and used a graph neural network (GNN) to predict mesh edge occupancy, constructed a dual graph on the faces. Other networks [191], [192] first converted point clouds to 3D voxels to take advantage of 3D CNN for reconstruction. However, a significant drawback of these approaches was their computational complexity. Predicting a dense mesh with numerous triangles was resource-intensive, especially for high-resolution point clouds.

To mitigate this complexity, alternative approaches use primitives. These methods start with basic primitives and then deform them to fit the input point cloud. Various types of primitives are used, such as square mesh [193], [194], local patches [195], sketch curves [196], cube mesh [197], box primitive [198], volumetric primitive [199], super-quadric primitive [200], convex primitive [197,198], and constructive solid geometry

(CSG) tree. For instance, in Point2Mesh [203] and Category-Specific Mesh Reconstruction (CMR) [204], where the initial primitive is a convex hull of the point, they treat the 3D reconstruction problem as a mesh optimization problem, iteratively refining the primitive by deforming its subcomponents. Unlike approaches that rely on pre-defined single primitive, some deep learning methods can directly identify the combination of various primitives present within a point cloud [205], [206], [207], [208], [209], [210]. Towards concise representations, CSG [211] is widely employed for shape modeling due to its capacity of multi-Boolean operations (e.g., union, intersection, and difference) to predict the combination of boxes, spheres, cylinders, and cones primitives [212], [213]. For instance, Paschalidou et al. [212] introduced to represent 3D model using a union operation of super quadrics as primitives. Chen et al. [201] and Deng et al. [202] proposed to first decompose space to a set of convex shapes, then use a two-stage network to learn a continuous approximation of meshes. These approaches can handle a broader range of object shapes compared to using a single primitive type, achieving compact representation by decomposing complex objects into simpler primitives.

*3.4.1.2. Implicit representation*

Implicit representations have emerged as a powerful approach for 3D mesh reconstruction. Unlike explicit methods that directly predict mesh structures, implicit methods learn a continuous function that maps a 3D point to a scalar field. This scalar field can represent various properties, such as: 1) a signed distance field (SDF) [214], [215], [216], [217] to indicate whether a point lies inside or outside a surface by its distance to the closest surface point, 2) a continuous decision boundary using an occupancy function [192], [218], [219], [220], [221] to define the probability of a point belonging to the object's interior; and 3) an unsigned distance field [191], [222], [223], [224] to represent the distance to the closest surface point. Once the implicit function is learned, mesh surfaces can be extracted by applying meshing or iso-surfacing algorithms like Marching Cubes (MC) [225] or Dual Contouring (DC) [226] to convert the continuous representation into a polygonal mesh.

Several methods directly learn the implicit function from a point cloud using neural networks, particularly MLPs. For example, Stucker et al. [227] reconstructed an iso-surface of a 3D scene from a continuous occupancy field using learned embeddings of photogrammetric point clouds and ortho-photo stereo pairs. Erler et al. [218] estimated an SDF with local and global feature vectors, demonstrating the strong generalization capabilities of implicit field learning. NDF [191] mapped point clouds to surfaces as a prior implicit model, predicting the unsigned distance field for both closed and non-closed 3D shapes.

Alternatively, some methods learn the implicit representation on 3D voxel grids. For instance, Chibane et al. [228] used convolutional encoders to capture multi-scale features from the voxel grid. These features were then used to predict inside-outside signs for query points. Mai et al. [229] employed scaled axis-aligned anisotropic 3D Gaussians to represent the implicit field on voxels. Park et al. [215] represented a shape's surface with a continuous volumetric SDF, and the learned representation implicitly encoded a shape's boundary as the zero-level set of the learned function while representing the classification of space as being part of the shape's interior or not. Mescheder et al. [230] implicitly represented 3D shapes as the continuous decision boundary of a deep network classifier and achieved relatively higher resolution and lower memory occupation. Wang et al. [231] leveraged octree and grid voxels to adaptively subdivide space for efficient complex shape representation.

Explicit approaches provide an intuitive understanding of the reconstructed surface and are well-suited for applications where mesh topology is crucial. Implicit representations, on the other hand, capture fine geometric details and offer flexibility in representing complex shapes. Despite their advantages, both methods face challenges such as handling noisy or incomplete point clouds and managing the complexity of reconstructed shapes.

*3.4.2 LoD Modeling*

LoD models [232] are a cornerstone of Geographic Information Systems (GIS), playing a critical role in spatial analysis [233]. As shown in Figure 4, these models represent buildings at various levels of detail, ranging from simple 2D footprints (LoD0) to highly detailed 3D structures with windows, doors, and roof features (LoD3). The ability to represent buildings at different levels of detail unlocks a vast array of applications. For example, LoD2 models, which include basic roof structures, can be used to estimate the energy demanded by buildings [234], building material stock [235]. For LoDs building reconstruction from point cloud data, point clouds are the go-to data source. However, these point clouds are typically noisy with a non-uniform point density, which can pose significant challenges in accurately reconstructing building surface geometry.

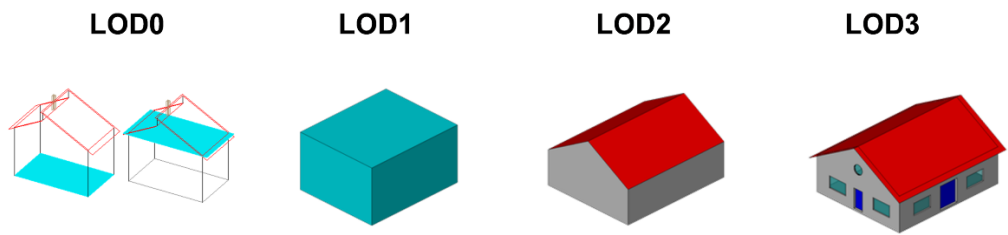

*Figure 4. The four representations of LoD defined by CityGML 3.0 for building exterior modeling. This figure is taken from the OGC CityGML 3.0 Conceptual Model Standard [232].*

**LoD0 models**, representing a building's footprint as a simple 2D polygon, form the essential foundation for spatial analysis in GIS [236]. These basic building outlines, typically obtained from sources like GIS Open Data (e.g., OpenStreetMap) or extracted from satellite imagery. Deep learning emerges as a valuable tool for enhancing existing LoD0 data. For example, deep learning models trained on aerial photogrammetric point clouds or aerial laser scanning point clouds can automatically detect and classify buildings [237], [238]. However, due to their 2D nature, those models lack information about building height and complex shapes, necessitating higher LoD models for applications requiring detailed building geometry [239].

**LoD1 models** provide basic volumetric representations for various applications. Traditional methods that use elevation information from point clouds can automatically extrude 3D LoD1 buildings from 2D building footprints [240]. This approach offers a fast and efficient way to generate basic 3D city models. Deep learning can further automate the process by reconstructing the LoD0 from airborne point clouds or aerial photogrammetry [241]. Then the extrusion-based method is applied to generate large-scale LoD1 models. For example, Teo [242] proposed a Fully Convolutional Network (FCN)-based method to detect initial building outlines from LiDAR data and automatically reconstruct 3D prismatic building models by assigning heights to the building outlines. However, extrusion-based LoD1 models have limitations, as they may not capture complex roof structures or building details.

**LoD2 models** represent buildings with a more detailed block shape and defined roof structures. To model roof structure, Li et al. [243] identified a set of candidate corner points from the deep features extracted by PointNet++ [119] and then predicts roof edges from an exhaustive set of candidate edges between the vertices. One significant challenge is the difficulty in accurately extracting the optimal corner points or edge primitives from low-quality LiDAR data in urban environments. To address the challenge of noisy and occluded point clouds, Zhao et al. [244] and Akwensi et al. [245] proposed to integrate point completion into a 3D reconstruction pipeline, allowing a more accurate geometric representation before reconstructing the building model. However, a fixed number of points of input in these networks can limit the representation of complex models, and it is hard to be applied to large-scale city scenes.

Given the complexity of 3D building models, some methods leverage additional data sources beyond point clouds to achieve a more detailed reconstruction. For example, primitives and holistic primitive fitting methods were used along with PointNet++ [119] for 3D building reconstruction from point clouds [246]. Kippers et al. [247] proposed a method that utilizes existing 3D BAG CityJSON city models [239] and floor plan images alongside point clouds for building reconstruction. Kada [248] used building footprints deduced from the boundaries of roof point sets, along with segmented roof faces and their slopes obtained from roof points, to reconstruct 3D building models. By incorporating various deep learning models and data sources, LoD2 reconstruction methods can achieve an accurate and detailed representation of building geometry, but the façades of these models are reconstructed by the assumption of the roof boundaries extracted from airborne point clouds.

**LoD3 models** are crucial for applications like evaluating the solar potential of buildings [249], since they provide a rich representation of buildings, including façades, windows, doors, and intricate roof structures. The data sources of reconstruction of LoD3 models are usually multi-images or Mobile Laser Scanning (MLS) to acquire the details of façades elements. Pantoja-Rosero et al. proposed combining LoD2 models with 2D image semantic segmentation information for openings to generate LoD3 models [250]. Recently, Scan2LoD3 [251] leverages the physics of laser rays to provide geometrical cues and incorporates existing 3D semantic building models to probabilistically identify potential model conflicts for refining existing 3D models. However, the end-to-end LoD3 generation remains challenging. The limitations include the need for co-registered data from both airborne and MLS point clouds, which may not always be available. Furthermore, uneven density, noise, and missing data around openings like windows further hinders the accuracy and automation of reconstruction algorithms.

Deep learning's power in LoD model reconstruction streamlines the process of creating 3D building representations in different details. While lower LoDs might be achievable with both conventional and deep learning approaches, LoD2 and LoD3 reconstruction benefit from additional attributes and data sources. The intricate complexity of building geometries presents challenges for deep learning models to automatically capture the fine details required for LoD2 and LoD3 reconstruction. Furthermore, scaling these high-detail modeling methods to the urban level often faces scalability and computational constraints. Memory-efficient approaches are critical for urban-scale modeling, for instance, Song et al. [252] highlighted the advantages of view-based methods for large-scale mesh reconstruction. Addressing these challenges demands robust algorithms that can adapt to varying spatial and geometric complexities, enabling reliable generations of high-detail models from multi-source point cloud data.

## 4. An overview of deep learning-based point cloud processing applications

*Table 2. A selection of applications using 3D point clouds*

| Applications | Volume of processed data | Difficulty of collecting data | Complexity of the models | Level of autonomy |
|---|---|---|---|---|
| City Modeling | +++ | +++ | +++ | ++ |
| Forestry | ++ | + | ++ | +++ |
| Agriculture | ++ | + | + | ++ |
| Ecology | +++ | + | + | ++ |
| Utility Mapping | + | +++ | ++ | + |

## 4.1. City Modeling

Deep learning has become a cornerstone for creating advanced city models. This integration with point cloud data enables sophisticated tasks such as segmentation, semantic segmentation (Section 3.3), and mesh or surface reconstruction (Section 3.4). A 3D city model is a digital representation of an urban environment, capturing the three-dimensional geometry of buildings, roads, and other essential elements [253]. These models serve as crucial tools for various applications in smart city development [254], [255]. This section provides a comprehensive overview of how deep learning models are utilized in different aspects of city modeling, emphasizing their role in propelling the development of smart cities in various applications.

*4.1.1 Digital infrastructure management*

Digital infrastructure management through digital twins is gaining increasing attention [256], [257]. 3D city models generated through deep learning and point cloud data offer a detailed and accurate representation of urban infrastructure, including buildings, roads, bridges, and underground utilities. This allows urban planners and engineers to visualize and assess the impact of infrastructure projects, monitor existing structures, and inform decision-making [254].

In utility management, deep learning techniques have revolutionized utility management by facilitating efficient identification and analysis of urban objects using point clouds. For example, roadside objects such as cars, trees, poles, pedestrians, bicycles, and vegetation are crucial components of the urban environment. Deep learning models can be trained to segment point clouds, separating and classifying these objects with point clouds [258]. Combining point clouds with data from other sources, such as multi-view images, can further enhance object recognition [259]. Some studies focus on specific roadside objects. For example, one approach uses a segmentation-then-classification method to extract pole-like objects [260] for use in road inventory studies [261], [262]. Another study leverages deep learning methods to recognize roadside plants for street safety management [263] as an instance recognition task [264]. Beyond roadside objects, deep learning models also demonstrate promising applications in extracting power lines [265], tunnels, and underground infrastructure [266] are also can be extracted by point cloud interpretation using deep learning models.

One of the most promising applications in infrastructure management is structural health monitoring. Deep learning algorithms can analyze 3D models to continuously monitor bridges [267] and buildings [268], [269], [270] for signs of deterioration such as cracks [268] and deformations [271]. The detected infrastructure allows for proactive maintenance, extending infrastructure lifespan, and reducing costs associated with reactive repairs. For instance, Wang et al. [272] employed a point-based network for detecting and evaluating the worn condition of pavement markings, highlighting the versatility of deep learning in addressing various infrastructure elements. Furthermore, leveraging deep learning models to analyze historical data and identify patterns makes predictive maintenance becomes possible to indicate potential failures [273], [274].

The interdependence between critical infrastructure systems creates a unified urban system, making decisions regarding infrastructure management for city planning more complex and requiring a thorough understanding of city models [275]. In urban planning, detailed 3D models derived from point clouds through deep learning can be integrated with GIS to provide spatial context and enable advanced analysis. This integration allows for spatial context analysis and informed decision-making when planning infrastructure projects within the broader cityscape [276]. For instance, Rojas-Rueda et al. [277] demonstrated this by analyzing green spaces within cities using deep learning and GIS, enabling informed urban planning decisions regarding these vital elements.

Despite significant advancements, seamlessly integrating various technologies like LiDAR, GIS, and deep learning into a cohesive workflow presents technical hurdles. Addressing these challenges through future research and advancements will be essential for unlocking the full potential of deep learning and 3D city modeling in digital infrastructure management.

*4.1.2 Energy Management*

The ever-increasing demand for energy necessitates a two-pronged approach: optimizing energy usage and integrating renewable sources. This is critical for fostering sustainable and resilient urban environments while tackling climate change challenges [278]. 3D city models generated from aerial lidar point clouds, particularly those with accurate roof structure information, play an important role in achieving these goals [279]. By integrating these models with energy data and deep learning-based analytics, urban planners and engineers can develop effective strategies to enhance energy efficiency and promote sustainability.

In the context of energy consumption analysis, deep learning empowers us to estimate and predict energy usage in buildings and urban areas. 3D city models, particularly LoD1 or LoD2 are integrated with deep learning techniques to analyze energy consumption patterns [280]. For instance, Fan et al. [281] introduced a CNN-based model to extract features for historical energy consumption data. This is then followed by a RNN-based architecture to predict short-term energy load of buildings. Gao et al. [282] proposed a sequence-to-sequence (seq2seq) model with transfer learning to forecast energy consumption with limited historical data for the predicted/target buildings. Some studies further integrate sensor data from Internet of Things (IoT) to enhance prediction accuracy and enable real-time application [283].

3D city models are also crucial for spatial analysis in the context of renewable energy integration. For example, by analyzing the urban landscape, these models help identify suitable locations for solar panel installations on rooftops and other structures. Nam [284] introduced deep learning to forecast the economic and environmental costs associated with different renewable energy scenarios, aiding optimal installation decisions in Korea. Zhou et al. [285] proposed using the location information of solar panels on rooftops or facades in building models to estimate the potential for zero-energy estimation at the city level. It is important to note that, in these methods, deep learning is not only used to generate 3D city models, but also to analyze and simulate energy scenarios based on these models.

The combination of city modeling and energy management fosters the development of sophisticated systems that enhance overall energy efficiency in urban settings. While point cloud generated LoD2 building models are widely used in these applications, offering significant insights into the sustainable energy practices needed in urban settings. However, they often lack essential building information in these automatically reconstructed models for precise energy demand and heating/cooling load estimation. For instance, the "year of construction" and "materials" of buildings are data points, but may not automatically capture in the reconstructed models and need to be manually incorporated into LoD2 models [286]. Furthermore, seamlessly integrating GIS analysis and deep learning methods remains a challenge that requires further research.

*4.1.3 Cultural Heritage*

In the field of cultural heritage, 3D point clouds are increasingly serving as the foundation for as-built BIM models [287], enabling the reconstruction of ancient sites and fostering a deeper understanding of our cultural heritage. One of the most significant contributions of deep learning lies in streamlining the creation of detailed 3D models for heritage sites. Pan et al. [288] proposed a "heritage digital twin platform" that integrates point clouds, semantic information generated by deep learning models, and high-fidelity 3D BIM. Similarly, Russa et al. [289] leveraged deep learning and advanced data processing to efficiently generate detailed 3D models, improving the accuracy and speed of cultural heritage documentation. These advancements are crucial for preserving historical sites and artifacts for future generations.

The advancement of deep learning based semantic segmentation [290] is particularly beneficial for cultural heritage reconstruction, as it allows researchers to automatically identify and categorize various features within a scanned site [287], [291]. While deep learning has significantly enhanced visualization-centric applications like 3D reconstruction, its use in non-visualization areas like structural analysis and predictive maintenance for cultural heritage sites remains limited since the complex and various geometries. Further research is needed to bridge this gap and unlock the full potential of deep learning for these crucial aspects of cultural heritage preservation.

*4.1.4 Road Network Mapping from Point Clouds*

Road networks are the backbone of urban transportation systems, playing a crucial role in facilitating urban planning and management. Point cloud generated through MLS [37], airborne laser scanning (ALS) [292], and terrestrial laser scanning (TLS) [293], [294] serve as reliable data sources for producing highly accurate, precise, and dense georeferenced 3D road networks [295], [296]. Complementing this, deep learning techniques have emerged as a powerful tool for automatic road network modeling, enabling efficient extraction of road information from point cloud data. For instance, Jiang et al. [297] utilized semantic segmentation networks to extract railway features from point clouds, facilitating accurate railway reconstruction. Similarly, Wang et al. [298] proposed a network to segment road surfaces and associated infrastructure from point clouds. By leveraging additional processing techniques, they were able to extract road boundaries and centerlines. In addition to road networks, deep learning methods are widely used for the automatic extraction of road markings [299], [300], and road signs by semantic segmentation task or object detection task [301].

Reconstructed road networks form the foundation for various traffic management applications. Traffic demand estimation, a cornerstone of modern transportation infrastructure and Intelligent Transportation Systems (ITS), benefits significantly from deep learning [302]. Due to the inherent network structure of road networks, Graph Neural Networks (GNNs) have become the preferred architecture for traffic demand estimation tasks. For instance, Zhao et al. [303] employed a GCN module to capture the spatial and temporal dependencies influencing bus travel demand. The short-term bus travel demand is then forecasted by fusing the dynamic built environment influences and spatiotemporal dependencies using a geographically weighted regression method. The Conjoint Spatio-Temporal graph neural network [304] constructs heterogeneous graphs from prior and posterior information to capture high-order spatio-temporal relationships. Additionally, a semantics-aware dynamic graph convolutional network (SDGCN) is proposed for traffic flow forecasting [305]. These studies demonstrate how GNNs exploit spatial features using convolutional operations, while incorporating temporal features through Gated Recurrent Units (GRUs).

GNNs effectively capture interactions between nearby traffic sensors or stations, thereby improving prediction performance. Despite the advancements, GNN-based models face challenges related to generalizability across diverse road networks, particularly between urban and rural areas. Furthermore, real-time traffic management applications necessitate fast processing speeds, and the inference time of GNNs requires further optimization.

*4.1.5 Disaster management*

Natural disasters like floods, wildfires, earthquakes, and hurricanes pose a significant threat to infrastructure and communities worldwide. Traditional methods for disaster monitoring and damage assessment often rely on laborious manual procedures, hindering the timely delivery of critical data for efficient response and recovery efforts [306]. Integrating deep learning techniques and 3D models into these processes offers a transformative approach for enhancing mitigation strategies and disaster recovery [307].

Timely and accurate monitoring is essential for early detection of potential hazards, enabling effective disaster preparedness and response. One innovative approach utilizes Unmanned Aerial Vehicles (UAVs) in conjunction with deep learning to assess the resilience of utility poles by automatically estimating their inclination angles, as demonstrated by Alam et al. [308], can automatically assess the resilience of poles. Such advancements exemplify how integrating deep learning with UAV technology can revolutionize disaster monitoring practices.

During disaster response, precise object detection and segmentation of point clouds are crucial for identifying vulnerable areas and optimizing evacuation routes. For example, Fang et al. [309] proposed a method for rapid flood risk mapping that integrates large-scale point cloud semantic segmentation with RandLA-Net [156] and hydrodynamics simulations, alongside hydrodynamics simulations. Vyron et al. [310] utilized deep learning to swiftly identify optimal landing spots for expedited disaster response operations.

In post-disaster scenarios, Xiu et al. [271] proposed the use of deep learning to detect collapsed buildings following earthquakes, aiding in search and rescue efforts. Liao et al. [311], [312], [313] further demonstrated the use of deep learning for classification, detection, and data acquisition from airborne point clouds to comprehensively assess disaster damage.

In risk-informed decision-making, 3D models are valuable for visualization in disaster management, however, their capabilities for supporting complex decision-making processes are limited. To enhance risk-informed decision-making, urban planners require more robust models that facilitate in-depth analysis and knowledge interpretation for effective disaster risk management and planning resilient urban environments.

From 3D city modeling to traffic management and disaster response, deep learning offers significant advantages over traditional geospatial modeling approaches. They streamline and expedite point cloud data processing and geoinformation systems workflows [314]. These deep learning approaches have demonstrated impressive performance in numerous applications that highlight the transformative potential of integrating point cloud and deep learning across various domains. However, these methods also encounter significant limitations. The quality and availability of data are paramount for ensuring the accuracy and completeness of model predictions. And the dependence on large training datasets can hinder generalization capabilities, especially when applied to different scenarios. Furthermore, the "black box" nature of deep learning models can hinder interpretability and trust, particularly in areas like reinforcement learning for smart city control systems. Lastly, while various methods exist for generating accurate semantic 3D city models to enable spatial and thematic analyses, the modeling process remains laborious and time-consuming [315]. 3D city models constructed from 2.5D datasets often lack crucial information present in detailed 3D point cloud data, relying instead on interpolated height data [275].

**4.2. Forestry**

The integration of point clouds with deep learning has revolutionized forestry management. For instance, object detection identifies individual trees and forest features, segmentation delineates these objects into meaningful clusters, and classification categorizes them into species or health classes. These capabilities are crucial for developing comprehensive forest models, which are essential for sustainable management and conservation strategies. Various architectures and algorithms play a pivotal role in processing the intricate spatial structures of forests, enabling precise monitoring and management actions that are specifically tailored to the unique needs of forest ecosystems.

*4.2.1. Tree segmentation*

LiDAR point clouds-based single tree segmentation is a key step to measure the structural parameters of a single tree by extracting point cloud data from remote sensing data of a complex scene, and segmenting a single tree crown from a point cloud is a rather difficult task. Point-based segmentation of single trees can effectively solve the problem of information loss when forming raster surfaces.

LiDAR point cloud-based single tree segmentation is a critical step in extracting structural parameters of individual trees from complex scenes. This process involves isolating single tree crowns from point cloud data derived from remote sensing, which presents significant challenges. Point-based segmentation of single trees can effectively address the issue of information loss that occurs when converting point clouds into raster surfaces.

Morsdorf et al. [316] initially identified local highest points of trees using a Digital Surface Model (DSM) and analyzed them via the K-means clustering algorithm, marking one of the first attempts to use point cloud data for individual tree segmentation. Subsequently, Wang et al. [317] advanced the segmentation technology by voxelizing point cloud data and partitioning the crown areas based on the elevation distribution of voxels, this method provided a more structured approach to segmenting individual trees, offering practical utility in assessing tree canopy structures and spatial distribution for forestry operations. Subsequently, Wang et al. [317] advanced this technology by voxelizing point cloud data and partitioning crown areas based on voxel elevation distribution. This method provided a more structured approach to segmenting individual trees, offering practical utility in assessing canopy structures and spatial distribution for forestry operations. Ayrey et al. [318] introduced a layer stacking algorithm that enhanced the precision and efficiency of segmentation and mapping forests through layered processing of point cloud data. Vega et al. [319] proposed extracting optimal vertex sets from different scales and combining them with K-means clustering, which proved useful for under-canopy management and regeneration efforts in forests. Dong et al. [320] utilized UAV remote sensing images combined with spectral enhancement techniques, integrating the DBI index to highlight fine tree features, thereby supporting applications such as monitoring tree health and assessing biodiversity. Furthermore, Cao et al. [321] introduced a deep learning algorithm based on YoloV7 in the semantic segmentation of trees, enhancing model performance in complex environments. Additionally, Liang et al. [322] proposed a novel tree energy loss algorithm that simulates low-level and high-level pairwise affinities by representing images as a minimum spanning tree. Recently, Shaheen et al. [323] proposed a self-supervised learning framework that eliminates the need for manual annotations by integrating transformation-invariant feature learning and an energy-based soft clustering mechanism. Their method demonstrated strong generalization on high-density LiDAR data, significantly reducing over-segmentation and outperforming traditional clustering techniques in both convexity and shape fidelity. In the context of UAV-based acquisition, Li et al. [324] introduced a region-growing method guided by density-adaptive canopy morphology priors, enabling the separation of overlapping crowns through 3D shape constraints and local density estimation, which proved especially effective in dense forest scenarios. Additionally, Lu et al. [325] developed ForestTreeSeg, a deep learning-based pipeline that leverages voxelized canopy representations and a two-stage 3D CNN to segment individual trees with high precision, outperforming existing rule-based and machine learning approaches in UAV-LiDAR datasets.

These studies demonstrate that with technological evolution, from simple cloud-based clustering to complex layered and optimized algorithms, individual tree segmentation technology has significantly improved in precision and efficiency, offering more accurate tools for forest management and monitoring.

These studies illustrate those technological advancements, from basic point cloud-based clustering to sophisticated multi-layered and optimized algorithms, have substantially enhanced the precision and efficiency of individual tree segmentation technology. This progress provides more accurate and reliable tools for forest management and monitoring.

*4.2.2. Tree species classification*

Accurate classification of tree species is a critical foundation for effective forest resource management, as its precision directly influences the decision-making process in forestry planning. The LiDAR system offers robust penetration capabilities, enabling it to pass through canopy gaps and extract characteristic parameters of individual tree structures both above and below the canopy.

In recent years, the application of deep learning technologies in tree species classification has achieved significant advancements. Mizoguchi et al. [326] improved classification accuracy by converting ground-based LiDAR-derived individual tree point clouds into depth maps and extracting features using CNNs, thereby enhancing the mapping of individual trees, which is crucial for monitoring growth patterns and assessing forest health. Blackburn et al. [327] introduced a scalable deep learning pipeline using eigenfeatures and 2D CNNs, achieving over 94% species classification accuracy from both UAV and airborne LiDAR, and validating the utility of low-density point clouds in complex forest environments. Guan et al. [328] utilized mobile LiDAR data, transforming point clouds into three-dimensional voxels and integrating deep learning techniques to improve classification accuracy, demonstrating the effectiveness of deep learning in analyzing dynamic forest environments. To address the challenges posed by complex forest areas, Zou et al. [329] employed voxelization and hierarchical feature extraction methods to better classify the tree species. These improvements are especially beneficial for dense or diverse forests, where species identification is often complicated by overlapping canopies and heterogeneous environments. For aerial perspectives, Briechle et al. [327] proposed an automated pipeline using eigenfeature-enhanced CNNs to classify seven tree species from UAV-LS and ALS data, achieving over 94% accuracy and demonstrating that low-density lidar data can support scalable species classification in complex forest environments. Briechle et al. [330] applied PointNet++ for deciduous forests, while Liu et al. [331] proposed LayerNet based on 3D deep learning, which extracts regional features and integrates global features from UAV LiDAR data. These methods enable resource managers to efficiently assess species distribution over extensive areas. To improve species classification in complex forests, Tockner et al. [332] proposed an interpretable method combining random forests with rule-based reasoning, achieving 89.8% accuracy across nine species and highlighting the importance of upper-canopy intensity and geometric ratios for explainable, trait-driven predictions. Dominik et al. [333] developed a method that projects three-dimensional point clouds into two-dimensional images and uses CNNs to extract image features for tree species classification. The results showed improvements in classification accuracy for ash, oak, and pine trees by 6%, 13%, and 14%, respectively, highlighting the potential of targeted species identification in informing reforestation efforts and biodiversity studies.

Additionally, Hell et al. [334] designed a new 3D-CNN structure that takes full-spectrum bands as input without the need for additional preprocessing or postprocessing. This structure fully exploits the spectral and spatial information of hyperspectral images, significantly reducing the demand for GPUs and enabling advanced tree species mapping in regions with limited computing resources. These studies not only enhance the accuracy of tree species classification but also reduce processing time, providing strong technical support for precise forest resource management.

**4.3. Agriculture**

Deep learning has significantly enhanced agricultural practices by leveraging point cloud data, making agricultural management and application more advanced. This includes identifying crops and agricultural equipment, classifying crop elements into different groups, and categorizing them based on crop type or health status. These tasks are crucial for promoting efficient monitoring and management of resources.

Guan et al. [335] utilized airborne 3D laser scanning technology and the mean shift algorithm to effectively enhance the accuracy of soybean canopy geometric parameter extraction, reducing the relative error from 9.05% to 5.14%. In more specialized applications, Gong L et al. [336] designed the Panicle-3D algorithm for rice panicle phenotypic parameters, based on 3D point cloud CNNs, achieving a high segmentation accuracy of 93.4% and an IoU of 86.1%. Finally, Zhu et al. [337] established the CropQuant-3D platform by integrating backpack LiDAR and 3D computer vision technologies for large-scale field phenotyping of wheat, further exploring the complex relationships between crop phenotypes and environmental factors. Additionally, Zhu et al. [338] developed a platform for collecting soybean canopy images using Kinect sensors, effectively calculating soybean plant height and leaf area index, and achieving rapid and efficient collection of point cloud data in outdoor environments. These technologies provide robust algorithmic support for field phenotyping analysis.

Guan et al. [335] employed airborne 3D laser scanning technology combined with the mean shift algorithm to significantly improve the accuracy of soybean canopy geometric parameter extraction, reducing the relative error from 9.05% to 5.14%. Qi et al. [339] summarized recent advances in 3D model-based phenotyping of grain crops across multiple scales, highlighting trait extraction in maize, wheat, rice, soybean, and sorghum, while addressing the integration of high-throughput platforms, advanced sensors, and AI in phenotypic data processing. In specialized applications, Gong et al. [336] introduced the Panicle-3D algorithm for rice panicle phenotypic parameter extraction using 3D point cloud CNNs, achieving a segmentation accuracy of 93.4% and an IoU of 86.1%. Zhu et al. [337] developed the CropQuant-3D platform by integrating backpack LiDAR and 3D computer vision technologies, enabling large-scale field phenotyping of wheat and facilitating the exploration of complex relationships between crop phenotypes and environmental factors. Ma et al. [340] presented a high-precision 3D reconstruction and phenotypic simulation method for soybean, integrating Kinect-based canopy imaging, DFSP segmentation, and the Richards growth model. It achieves over 96% accuracy in trait estimation and robust prediction of plant development over time. Furthermore, Zhu et al. [338] created a system for capturing soybean canopy images using Kinect sensors, which efficiently calculates plant height and leaf area index while ensuring rapid and accurate collection of point cloud data in outdoor environments. These advancements provide robust algorithmic support for field phenotyping analysis.

Furthermore, Wang et al. [341] introduced a method for 3D reconstruction of soybean canopies using multi-camera system. This approach involved acquiring comprehensive point cloud data from three distinct perspectives and preprocessing it with conditional filtering and k-nearest neighbors (KNN) algorithms. They utilized random sample consensus (RANSAC) and iterative closest point (ICP) algorithms for point cloud registration and fusion, thereby achieving accurate 3D reconstructions of the soybean canopy and enabling quantitative analysis of phenotypic traits such as plant height, leafstalk angle, and canopy width. Liang et al. [346] presented a 3D point cloud-based method for segmenting stems and leaves of tomato seedlings and extracting phenotypic traits, achieving high accuracy across multiple parameters and demonstrating its potential for robust, high-throughput phenotyping in agricultural breeding and cultivation systems Li et al. [343] tackled the challenge of internal canopy leaves being occluded by external branches in cotton plants by developing the Fragmented Leaf Point Cloud Reconstruction Algorithm (FLPRA). By integrating instance segmentation networks (ISN), GANs, and point cloud reconstruction algorithms (PRAs), they successfully reconstructed a 3D model of the cotton plant that included both internal and external canopy structures, effectively addressing the issue of incomplete point clouds due to external leaf occlusion. Additionally, Wu et al. [345] developed Plant-Denoising-Net (PDN), which leverages learning from the density gradient field of point clouds to enhance the accuracy of point cloud data processing, providing clearer and more precise data for plant phenotyping measurements. The advancements in these technologies not only resolve occlusion and noise issues inherent in traditional crop monitoring methods but also offer robust support for precision agriculture.

### 4.4. Ecology

The emergence of point cloud technology and recent advancements in deep learning algorithms enable the reconstruction of 3D structures from spatial variations in point density, thereby facilitating the inference of ecosystem characteristics. Deep learning models significantly enhance the accuracy of ecological assessments compared to traditional methods by processing large volumes of data rapidly and providing detailed classifications and volumetric analyses. These capabilities are crucial for monitoring changes in habitats and species distributions, which in turn supports the assessment of biodiversity and ecosystem health. Consequently, this approach facilitates proactive environmental management.

For instance, Xi et al. [347] employed various machine learning and deep learning classifiers for wood filtering and tree species classification, effectively identifying key areas of species characteristics. And Van den Broeck et al. [348] released a tropical-tree TLS dataset with manual leaf–wood labels, benchmarked state-of-the-art point-cloud networks within a unified pipeline, analyzed how segmentation quality influences woody-volume reconstruction, and openly shared their data, code, and trained models. Moreover, Ni et al. [349] developed a 3D segmentation framework that accurately extracts berry cluster features via 2D to 3D mapping, thereby enhancing the monitoring of fruit development and improving yield estimation. In forest ecology, Gonçalves et al. [350] designed a hybrid model integrating convolutional autoencoders with Partial Least Squares regression, successfully uncovering correlations between forest structure gradients and topographical changes in the Amazon forest. Patel et al. [351] demonstrated the potential of 3D deep learning models in precisely segmenting sorghum plant organs and extracting phenotypic traits, providing robust technological support for automating crop phenotyping tasks. Elias et al. [352] utilized deep learning to detect invasive species in both 2D and 3D spaces, effectively predicting their environmental spread using geospatial Long Short-Term Memory (LSTM) models, significantly reducing manual labor requirements. Krisanski et al. [353] further illustrated how deep learning can semantically segment high-resolution forest point clouds collected from diverse sensing systems and extract Digital Terrain Models (DTM) from these segmented point clouds, which are essential for forestry and ecological conservation. Luz et al. [354] developed a YOLOv8-based framework augmented with synthetic images for the early detection of invasive sun coral in underwater imagery, demonstrating rapid and reliable performance that underscores deep learning's effectiveness in bio-invasion monitoring. Additionally, Ma et al. [355] introduced the Forest-PointNet model, which focuses on precise segmentation of vertical structures in complex forest scenes. By integrating spatial and shape features of point clouds, this model accurately distinguishes between ground, shrubs, trunks, and leaves, thereby effectively reconstructing three-dimensional forest information to enhance our understanding of forest ecosystem functions. However, broader application of deep learning in ecology faces several challenges. The need for large, annotated datasets, the computational expense associated with processing 3D data, and the difficulty in generalizing models across diverse ecological settings pose significant hurdles. Further research is essential to overcome these challenges, ensuring that deep learning can fully support ecological conservation and research efforts.

### 4.5. Utility Mapping

Deep learning significantly enhances utility mapping by integrating point cloud data, enabling sophisticated tasks such as segmentation and 3D reconstruction of utility infrastructures. These 3D models digitally represent both subterranean and surface systems, which are crucial for maintenance and planning in urban management [356]. Liu et al. [357] introduced the Context-Aware Network (CAN), combining Local Feature Aggregation Modules (LFAM) and Global Context Aggregation Modules (GCAM). And Lu et al. [358] proposed Sen-net, a forest-tailored point-cloud segmentation that fuses spatial context, semantic detail, and adaptive guidance, achieving state-of-the-art accuracy on Lin3D and FOR-instance datasets. This approach not only preserves geometric details but also improves the network's robustness to noise, thereby enhancing the quality of semantic segmentation for large-scale urban point clouds. In road management, Soilán et al. [359] developed an automated method using the Point Transformer architecture to extract road markings from 3D point clouds and parameterize road alignments, providing substantial support for the digital management of urban transportation infrastructure. Additionally, Hansen et al. [360] launched the OpenTrench3D project, which collects public 3D point cloud data from various regions and employs advanced semantic segmentation models to detect and document underground utilities, offering new directions for transfer learning methods. Stranner et al. [361] utilized a 3D fitting algorithm to update existing utility line maps, accurately reflecting the actual built state, especially suitable for field use, and providing real-time, efficient technological support for urban planning and management. Zhou et al. [362] introduced TransPCNet, a Transformer-based point cloud classification network that detects sewer defects by analyzing 3D point cloud data. This model significantly improves 3D point cloud classification performance by enhancing feature interconnections. Additionally, Comesaña-Cebral et al. [363] introduced the ROADSENSE simulator, which generates labeled synthetic point cloud datasets using PointNet++ . This tool provides robust support for deep learning applications in urban traffic scenarios, thereby enhancing road infrastructure management. Meanwhile, Wei et al. [364] developed an automated algorithm for detecting and monitoring shape deformations in steel beam structures based on image-based 3D reconstruction (AREAS). Their method achieves high-precision structural health monitoring with an average error of less than 1 mm, highlighting the potential of deep learning in urban infrastructure management. Choi et al. [365] integrated ultrasonic tomography with image-based 3D reconstruction to effectively characterize internal damages in reinforced concrete columns. By combining P-wave velocity with external 3D point cloud data, this approach offers a comprehensive perspective for damage assessment. Wojtczak [366] devised a signal-energy-based method for identifying ultrasonic wave attenuation in heterogeneous materials, experimentally and numerically calibrating the Rayleigh damping coefficient; the technique streamlines model parameterization and enables early fracture detection in concrete.  Building on this, Fang et al. [367] expanded the application scope by developing an automated sewer inspection system that integrates inspection robots, SLAM, and instance segmentation techniques. This system not only improves inspection efficiency but also accurately locates damaged areas through 3D reconstruction, significantly reducing labor costs and risks. Besides, Xue et al. [368] introduced a SfM-deep learning method for tunnel leakage detection, demonstrating how enhanced brightness images and high-quality 3D texture models can reconstruct detailed three-dimensional structures of tunnels and precisely map and quantify leakage areas. Jensen et al. [369] introduced the BioVista dataset and showed that fusing 2D orthophotos with 3D airborne-laser-scanning point clouds via deep neural networks raised forest-biodiversity-potential classification accuracy, highlighting the complementarity of spectral and structural cues. Deep learning algorithms, particularly 3D CNNs, offer superior alternatives to traditional mapping methods by automating data processing and increasing accuracy and efficiency in detecting and cataloging utility assets. However, challenges such as data scarcity, high computational costs, and model generalization across different environments remain significant.

## 5. Discussions, Conclusions and Outlook

In this review, we have provided a thorough synthesis for point cloud data processing algorithms and the relevant broad range of tasks and applications. We first present different types of point cloud data collection methods, and then we focus our review on three major tasks, i.e., point cloud registration, scene completion, semantic interpretation and geometric modeling. Finally, we review a broad array of point cloud-based applications under the context of geospatial engineering.

The review shows that the progressive development of both point cloud data collection and processing methods are broadening the geospatial industry, encapsulating applications that were previously challenging, such as in forestry, agriculture and autonomous driving etc. While most of the existing works show that the ever-updating network architecture and increased volume of shared data for training in deep learning have achieved extra percents of performance improvement over benchmark datasets, these works, however, are not reported with consistent results when being applied to unseen datasets in practical applications. We show that despite the diversity of point cloud tasks, the process of feature extraction mostly follows voxel-based, view-based and point-based methods, the use of which depends on the volume of point clouds, size of the scene and the complexity of the tasks. This level of concept homogeneity allows researchers and developers to utilize new algorithms and concepts across different tasks, especially those that are closely related, such as semantic, instance, panoptic segmentation and object detection using point clouds.

There have been increasing attempts on the use of deep learning-based point cloud processing through various applications such as in city modeling, forestry, agriculture, ecology, utilize mapping etc. Many of the surveyed studies exhibit either first attempts or works that aim to improve on the use of DL, yielding elevated accuracy and efficiency over past practices. Yet, during the course of review, we found that the generalization capability of these methods, as well as their application to data at the geospatial volume and efficiency are still under-reported. Despite that there are many studies on the transferability of DL methods at the theory level and validated on typical benchmark dataset, scaling up these methods to the operational flows for use by practitioners remain challenging. Hence, future studies and novel algorithms on the point cloud processing can be more informative, if applied and validated on datasets in different domain applications at a geospatial scale.